%% file: Reprogramming_Survey_AAAI2024.tex
\documentclass[letterpaper]{article} % DO NOT CHANGE THIS
\usepackage{aaai24}  % DO NOT CHANGE THIS
\usepackage{times}  % DO NOT CHANGE THIS
\usepackage{helvet}  % DO NOT CHANGE THIS
\usepackage{courier}  % DO NOT CHANGE THIS
\usepackage[hyphens]{url}  % DO NOT CHANGE THIS
\usepackage{graphicx} % DO NOT CHANGE THIS
\usepackage{natbib}  % DO NOT CHANGE THIS AND DO NOT ADD ANY OPTIONS TO IT
\usepackage{caption} % DO NOT CHANGE THIS AND DO NOT ADD ANY OPTIONS TO IT
\usepackage{algorithm}
\usepackage{algorithmic}
\usepackage{amsmath,amssymb}
\usepackage{amsthm}
\usepackage{booktabs}
\usepackage{algorithm}
\usepackage{algorithmic}
\usepackage{color}
\usepackage{enumitem}
\usepackage{adjustbox}
\include{CPY_notation_V1}

\usepackage{newfloat}
\usepackage{listings}
\DeclareCaptionStyle{ruled}{labelfont=normalfont,labelsep=colon,strut=off} % DO NOT CHANGE THIS
\floatstyle{ruled}
\newfloat{listing}{tb}{lst}{}
\floatname{listing}{Listing}
\nocopyright
\title{Model Reprogramming: Resource-Efficient Cross-Domain Machine Learning}
\author{
Pin-Yu Chen\textsuperscript{\rm 1}
}
\affiliations{
        \textsuperscript{\rm 1}IBM Research\\
        pin-yu.chen@ibm.com
}

\usepackage{bibentry}
\begin{document}

\maketitle

\begin{abstract}
In data-rich domains such as vision, language, and speech, deep learning prevails to deliver high-performance task-specific models and can even learn general task-agnostic representations for efficient finetuning to downstream tasks. However, deep learning in resource-limited domains still faces multiple challenges including (i) limited data, (ii) constrained model development cost, and (iii) lack of adequate pre-trained models for effective finetuning. This paper provides an overview of \textit{model reprogramming} to bridge this gap. Model reprogramming enables resource-efficient cross-domain machine learning by repurposing and reusing a well-developed pre-trained model from a source domain to solve tasks in a target domain \textit{without} model finetuning, where the source and target domains can be vastly different. In many applications, model reprogramming outperforms transfer learning and training from scratch. This paper elucidates the methodology of model reprogramming, summarizes existing use cases, provides a theoretical explanation of the success of model reprogramming, and concludes with a discussion on open-ended research questions and opportunities. 
A list of model reprogramming studies is actively maintained and updated at \textcolor{blue}{\url{https://github.com/IBM/model-reprogramming}}.
\end{abstract}

\begin{figure}[t]
    \centering
    \includegraphics[width=1\columnwidth]{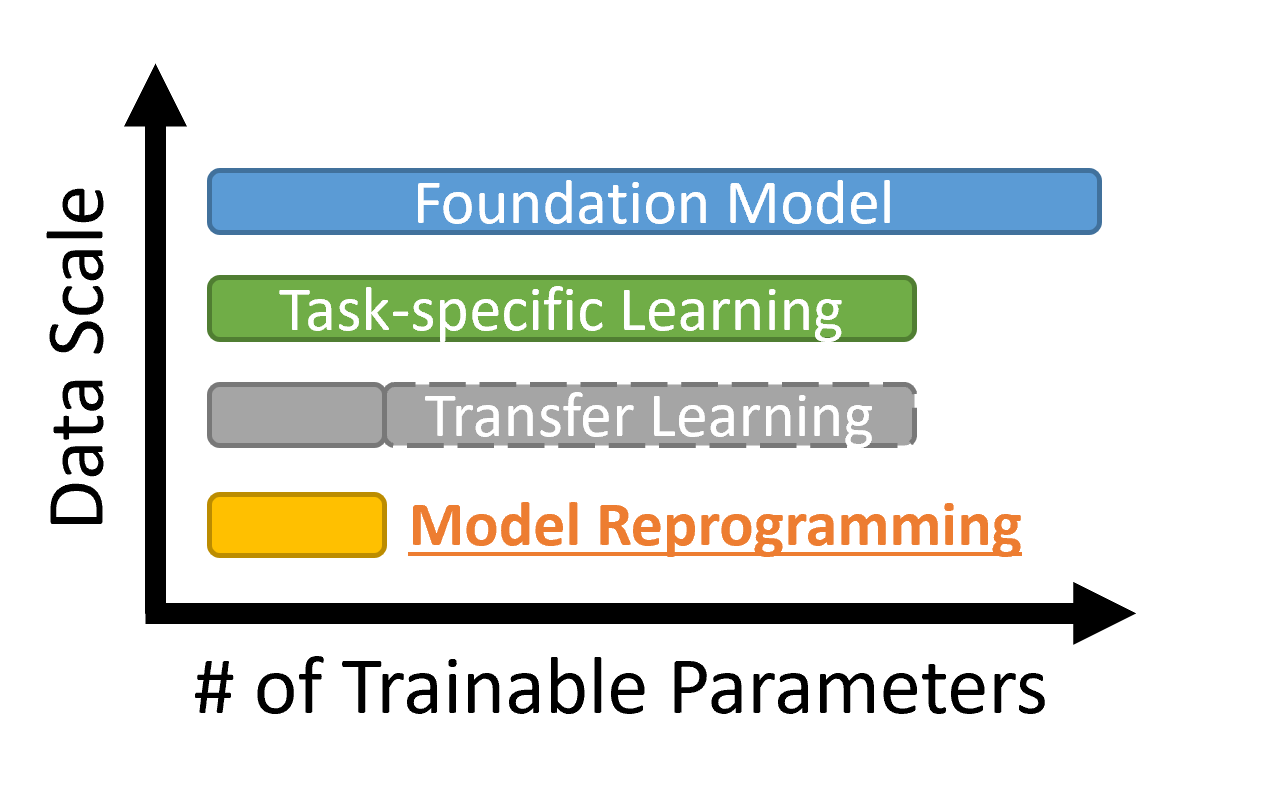}
    \vspace{-8mm}
    \caption{Visual illustration of data scale (bottom to top: small to large) and number (\#) of trainable parameters (left to right: small to large) in different machine learning paradigms. We note that the visualization does not reflect the actual relative differences due to excessively varying orders. A foundation model like GPT-3 has 175 billion trainable parameters and 499 billion tokens as training data. The trainable parameters in model reprogramming can be as few as the size of the data input (e.g., the number of image pixels can be in the order of thousands or fewer), and model reprogramming is particularly suited to small-scale data regime.
    In model reprogramming, the visualization does not take into account the pre-trained source model because it is kept intact and unchanged.
    The dashed box in transfer learning means variations in the number of model parameters used for fine-tuning, ranging from only training the last dense layer (linear head) to fine-tuning all parameters. The number of training epochs may also vary for each paradigm.  }
    \label{fig:illustration}
    \vspace{-4mm}
\end{figure}

\section{Introduction}

Designing and developing a top-notch deep learning model is a time-consuming and costly process. It is no secret that employing a high-capacity neural network model consisting of a tremendous number of trainable parameters, together with a proper selection of the network architecture and hyperparameter optimization, can lead to state-of-the-art machine learning performance when trained on a massive amount of data. Take the Generative Pre-trained Transformer 3 (GPT-3) \cite{brown2020language} as an example, which is one of the largest language models ever trained to date. GPT-3 has 175 billion parameters and is trained on a dataset consisting of 499 Billion tokens. The estimated training cost is about 4.6 Million US dollars even with the lowest priced GPU cloud on the market in 2020\footnote{See \url{https://lambdalabs.com/blog/demystifying-gpt-3}}. Such a large-scale language model is shown to be effective when applied to several downstream language-related tasks in \textit{the same domain (source)}. However, having invested so much to obtain a top-notch model, one interesting question to ask is: Can we reuse this valuable asset for machine learning in \textit{another domain (target)}, especially in the resource-limited setting when at least one of the following scenarios is concerned: (i)  lack of high-quality pre-trained models in the target domain for finetuning, (ii) scarcity of the available data in the target domain, and (iii) constraints on the model development and training cost (e.g. limited memory or training epochs).

To address these challenges, this paper provides an overview of the \textit{model reprogramming} framework towards resource-efficient cross-domain machine learning. The general rationale behind model reprogramming lies in repurposing and reusing a well-developed pre-trained model from a source domain to solve new tasks in a target domain \textit{without model finetuning} (i.e., model parameters are \textit{frozen}). Specifically, model reprogramming introduces an input transformation layer and an output mapping layer to the pre-trained source model to empower cross-domain machine learning.
Model reprogramming is favorable to the resource-limited setting because it (i) enables the reuse of pre-trained models from data-rich and well-studied domains (e.g., vision, language, and speech); (ii) attains data efficiency by only training the added input transformation and output mapping layers; and (iii) reuses available models and spares model development from scratch. The underlying working mechanism of model reprogramming in terms of the \textit{how} and the \textit{why}
will be explained in detail throughout this paper.

As a visual illustration, Figure \ref{fig:illustration} compares the data scale and the number of trainable parameters for different machine learning paradigms, including model reprogramming, transfer learning, task-specific learning (training from scratch), and foundation model (pre-training and fine-tuning). Data scale refers to the training data size. The number of trainable parameters relates to the model development cost, and models with more training parameters 
usually have higher training complexity.
We elucidate each paradigm as follows.
\begin{itemize}[leftmargin=*]
    \item \textbf{Model reprogramming} only requires training the inserted input transformation and output mapping layers while keeping the source pre-trained model intact (see Figure \ref{fig:reprog}). Notably, the number of trainable parameters does not include the source pre-trained model because its parameters are unchanged during model reprogramming. Therefore, it is particularly applicable to the small-data regime and features relatively low training complexity.
    
    \item \textbf{Transfer learning} is a common practice for in-domain knowledge transfer (e.g., pre-trained on an English-based task and finetuned for other English-related natural language processing tasks). The general principle is that some features learned from a source domain can be useful for machine learning in a target domain via model finetuning.
Transfer learning starts from a pre-trained source model and then finetunes a subset of the model parameters using the target-domain data. The number of trainable parameters can vary based on the size of the selected subset for finetuning.  If all model parameters are used for finetuning, the number of trainable parameters is the same as that of training from scratch, though the training epochs of transfer learning may be fewer. If only the last dense layer (i.e., a linear head) in the neural network is randomly initialized and made trainable, then the number of trainable parameters can be comparable to that of model reprogramming. The dashed box in Figure \ref{fig:illustration} indicates the variation in the size of trainable parameters for transfer learning. One notable limitation of transfer learning is that in some target domains, there may lack of adequate pre-trained models from similar domains for effective finetuning.

    \item \textbf{Task-specific learning} refers to training the parameters of a machine learning model by minimizing a task-specific loss (e.g., cross entropy classification loss on a given task). The model parameters are often randomly initialized and trained from scratch. In the small-scale data regime, training from scratch with large models usually yields unsatisfactory performance even when data augmentation is used.
    
    \item \textbf{Foundation model} \cite{bommasani2021opportunities}  features
    task-agnostic pre-training (often on a large-scale dataset) and efficient finetuning to downstream tasks. It is becoming a new trend in machine learning research due to its capability to learn general and discriminative representations of the considered data modality. The training of foundation model often follows the methodology of self-supervised learning, such as contrastive learning with self-generated positive and negative pairs, or masked token prediction. See \cite{jaiswal2021survey} for more details. The current practice of training foundation models still requires a massive amount of data for pre-training and a gigantic model for learning general-purpose representations, such as the GPT-3 model \cite{brown2020language}.
\end{itemize}

\begin{table}[t]
\centering
%\vspace{-2mm}
\begin{adjustbox}{width=0.99\columnwidth}
\begin{tabular}{@{}l|l@{}}
\toprule
Symbol                                                                                                    & Meaning                                                                   \\ \midrule
$\cS$ / $\cT$                                                                                             & source/target domain                                                    \\
$\cX_{\cS}$ / $\cX_{\cT}$                                                                                & the space of source/target data samples                                 \\
$\cY_{\cS}$ / $\cY_{\cT}$                                                                                 & the space of source/target data labels                                  \\
$\cD_{\cS}$ $\subseteq \cX_{\cS} \times \cY_{\cS} $ / $\cD_{\cT}$ $\subseteq \cX_{\cT} \times \cY_{\cT} $ & source/target data distribution                                         \\
$(x,y) \sim \cD$                                                                                          & data sample $x$ and one-hot coded label $y$ drawn from $\cD$              \\
$K_\cS / K_\cT$                                                                                                       & number of source/target labels                                                   \\
$f_\cS: \bbR^d \mapsto [0,1]^K $                                                                          & pre-trained  $K$-way source classification model                          \\
$\eta: \bbR^K \mapsto [0,1]^K$                                                                            & softmax function in neural network, and $\sum_{k=1}^K [\eta(\cdot)]_k =1$ \\
$z(\cdot) \in \bbR^{K}$                                                                                   & logit (pre-softmax) representation, and $f(x)=\eta(z(x))$                 \\
$\ell(x,y) \triangleq \|f(x)-y\|_2$                                                                                 & risk function of $(x,y)$ based on classifier $f$                          \\
$\bbE_{\cD} [\ell (x,y)] \triangleq \bbE_{(x,y) \sim \cD} [\ell (x,y)]$               & population risk based on classifier $f$                                   \\ 
$\delta$             & additive input transformation on target data            \\
 $\theta$~/~$\omega$ & parameters of input transformation / output mapping layers     \\    
  $M \in \{0,1\}^{d}$ & binary mask indicating which input dimension is trainable \\
    $h: \cY_\cS \mapsto \cY_\cT$ & source-target label mapping function
  \\    
\bottomrule
\end{tabular}
\end{adjustbox}
\caption{Mathematical notation}
\label{tab:notation}
\vspace{-4mm}
\end{table}

It is worth noting that the origin of model reprogramming can be traced back to the adversarial reprogramming method proposed in \cite{elsayed2018adversarial}. The authors in \cite{elsayed2018adversarial} demonstrate that ImageNet-1K image classifiers can be reprogrammed for classifying CIFAR-10 and MNIST images, as well as counting squares in an image, with mediocre accuracy. It is originally cast as an adversarial machine learning technique due to the implication that an attacker can leverage model reprogramming to repurpose the function of a pre-trained model without notice of the model provider, therefore causing ethical concerns or negative impacts. Beyond the adversarial purpose, the subsequent works such as \cite{tsai2020transfer,vinod2023reprogramming,yang2021voice2series} show that model reprogramming can be used as a resource-efficient cross-domain machine learning tool in a variety of problem domains and data modalities.

The remainder of this paper is organized as follows. Section \ref{sec_method} introduces the general framework of model reprogramming and summarizes existing works and use cases. Section \ref{sec_theory} provides theoretical explanations on the success of model reprogramming. Finally, Section \ref{sec_discuss} discusses some important open-ended research questions and opportunities for model reprogramming. For clarity, Table \ref{tab:notation} presents the main mathematical notations used in this paper.

\section{Model Reprogramming Framework}
\label{sec_method}

In this section, we start by introducing a generic framework and algorithmic procedure for model reprogramming (Section \ref{sub_method}). Then, we highlight some use cases of model reprogramming in current studies (Section \ref{sub_example}).

\subsection{Methodology of model reprogramming}
\label{sub_method}
Figure \ref{fig:illustration} shows a generic framework of model reprogramming and some examples of cross-domain reprogramming in the literature. In general, two new modules, an \textit{input transformation layer} and an \textit{output mapping layer}, are added to a frozen pre-trained source model to enable reprogramming. We will elucidate how these two layers are realized and implemented in the following paragraphs.

For ease of illustration, we focus on the setting of reprogramming a pre-trained classifier $f_\cS(\cdot)$ from a source domain to solve a classification task in a target domain. The notation of the model parameters associated with  $f_\cS(\cdot)$ is omitted because these parameters are fixed and unchanged during reprogramming. Without loss of generality, we assume $f_\cS(\cdot)$ takes a vectorized input of dimension $d_\cS$, which includes continuous data domains (e.g., image pixels, audio signals, numeric tabular features, etc) and discrete tokenized data domains (e.g., words, image patches, etc). The output of $f_\cS(\cdot) \in [0,1]^{K_\cS}$ is a ${K_\cS}$-dimensional vector of class prediction scores over ${K_\cS}$ source classes. 

We also note that the framework of model reprogramming makes no constraints on the source and target domains (e.g., domain similarity, knowledge transfer, etc), as long as their data formats are consistent. For instance, \cite{neekhara2022cross} shows an example of cross-domain reprogramming on an image classifier for sentence sentiment classification 
by mapping word tokens to image patches. Given consistent data formats, the only assumptions imposed by reprogramming are that (i) the target data dimension is no greater than the source data dimension (i.e., $d_\cT \leq d_\cS$); and (ii) the number of target class labels is no greater than that of source class labels  (i.e., $K_\cT \leq K_\cS$). These two assumptions are based on the rationale that model reprogramming is applicable to the setting of reprogramming a pre-trained model from a more complex source domain to a simpler target domain with a lower input dimension and a smaller number of class labels, but the reverse setting may not be applicable.

\begin{figure}[t]
    \centering
    \includegraphics[width=0.95\columnwidth]{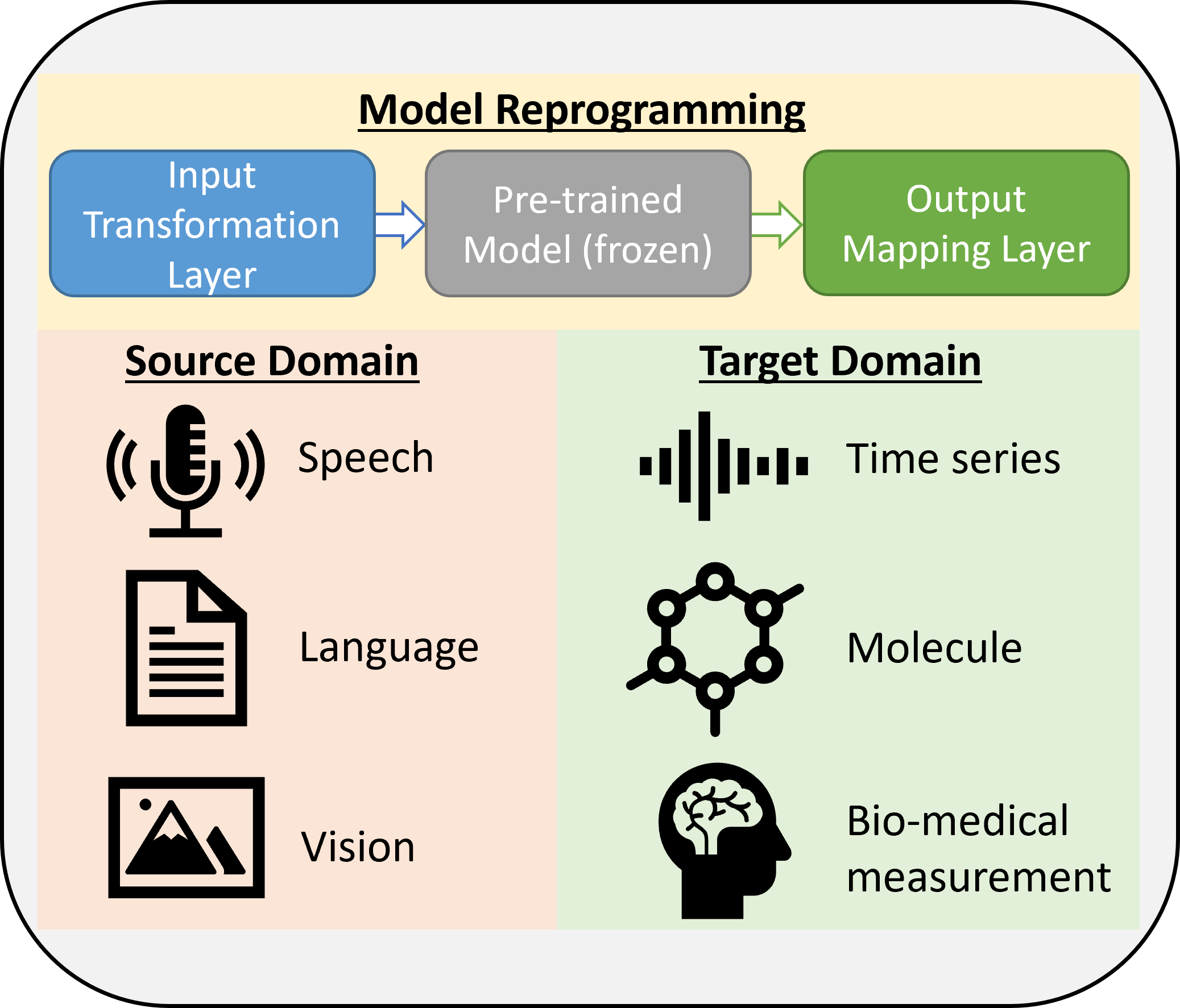}
    \caption{Illustration of the model reprogramming framework (top) and some examples of cross-domain machine learning via model reprogramming (bottom). Model reprogramming enables cross-domain machine learning by adding two modules, an input transformation layer (blue box) and an output mapping layer (green box), to a pre-trained model selected from a source domain.  When reprogrammed to solve target-domain tasks,
    the pre-trained source model is frozen and its model parameters are unchanged. Examples of cross-domain machine learning include reprogramming speech models for time-series \protect\cite{yang2021voice2series}, language models for molecules \protect\cite{vinod2023reprogramming}, and general imaging models for bio-medical measurements \protect\cite{tsai2020transfer}. }
    \label{fig:reprog}
        \vspace{-2mm}
\end{figure}

\paragraph{Input transformation layer.}
Given a target-domain data sample $x_\cT \in \bbR^{d_\cT}$, the input transformation layer transforms $x_\cT$
into another data sample $\xt_\cT \in \bbR^{d_\cS}$, which fits the input dimension of the pre-trained source model $f_\cS$. The input transformation function is parameterized with a set of trainable parameters $\theta$, which is formally defined as $\xt_{\cT}=\textsf{Input-Transform}(x_\cT|\theta)$. For continuous data, the transformation function can be as simple as a universal trainable additive input perturbation (i.e., a bias term) $\delta$ with a binary mask $M \in \{0,1\}^{d_\cS}$ indicating which dimension is trainable. Putting in mathematical expressions, we have 
\begin{align}
    \label{eqn_input_transform}
    \xt_\cT = \textsf{Zero-Padding}(x_\cT) + \underbrace{M \odot \delta}_{:= \theta},
\end{align}
where the operation \textsf{Zero-Padding}$(\cdot) \in \bbR^{d_\cS}$ means augmenting the input with extra dimensions of zero values, and the notation $\odot$ denotes the Hadamard (element-wise) vector product. The number of ones in $M$ indicates the location and the effective number of trainable parameters. For instance, in \cite{elsayed2018adversarial,tsai2020transfer}, a target sample $x_\cT$ is placed at the center of its zero-padded version $\xt_\cT$, where the location of $x_\cT$ is indicated by the masked index set $\{i: M_i = 0\}$ and hence by \eqref{eqn_input_transform} the masked set is not trainable. The remaining zero-padded dimensions, indicated by $\{i: M_i = 1\}$, are made trainable. We also note that if the data input range of $f_\cS$ is bounded, for example within $[-1,1]^{d_\cS}$, one can apply the change-of-variable technique to satisfy the constraint, such as making $\theta = \tanh(M \odot W)$,  where $W \in \bbR^{d_\cS}$ denotes unconstrained optimization variables and $\tanh$ denotes the hyperbolic tangent function. The input transformation function also allows placing multiple replicates of a target sample in the transformed data sample, such as in \cite{yang2021voice2series}.

For discrete data, the input transformation function can be a set of inserted trainable tokens at the data input in \cite{hambardzumyan2021warp}, or a trainable token embedding mapping function $\cV_{\cT} \approx \cV_{\cS} \theta$ via dictionary learning in \cite{vinod2023reprogramming}, where
the rows of $\cV_{\cS}/\cV_{\cT}$ are the embeddings of the source/target tokens.

\begin{table*}[t]
\begin{adjustbox}{max width=0.99\textwidth}
\begin{tabular}{@{}lllll@{}}
\toprule
Reference & Source domain & Source model & Target domain & Highlights \\ \midrule
\cite{elsayed2018adversarial}      &    General image        &     ImageNet        &  CIFAR-10/MNIST/counting           &  first work; mediocre accuracy          \\
\cite{neekhara2018adversarial}      &     Text       &  LSTM/CNN           &   Character/Word level tasks        &  context-based vocabulary mapping          \\
\cite{tsai2020transfer}      &  General image          &   ImageNet/API         &    Bio-medical measurement/image      &   black-box reprogramming; new SOTA       \\
\cite{vinod2023reprogramming}     &  Text         &   BERT         &    Biochemical sequence      &   vocabulary embedding mapping       \\
\cite{kloberdanz2021improved}       &   General image          &   ImageNet           &    Caltech 101/256 (reduced)          &   trainable input \& output layers         \\ 
\cite{lee2020reprogramming,dinh2022improved} &    Image/Spectrogram        &      GAN      & Image/Spectrogram           &     reprogram GAN to conditional GAN      \\ 
\cite{randazzo2021adversarial}    &   MNIST/lizard pattern       &   Neural CA     &   MNIST/Lizard pattern      &  stable out-of-training configurations   \\ 
\cite{hambardzumyan2021warp}      &  Text        &    BERT \& variants        &   GLUE/SuperGLUE      &     trainable tokens and data efficiency  \\ 
\cite{yang2021voice2series}      &   Speech          &      Attention-RNN      &   Univariate time series     &    new/same SOTA on 19/30 datasets        \\ 
\cite{yen2021study}  &   Speech          &      Attention-RNN      &   Low-resource speech     &   new SOTA; reprogramming+finetuning \\

\cite{chen2021adversarial} &   General image          &      ImageNet    &   Financial transaction     &   overlay image and transaction feature   \\
\cite{neekhara2022cross}  &    General image        &  ViT/ImageNet          &  Sequence      & text sentences and DNA sequences    \\
\cite{jing2023deep}  &    Graph        &  GNN       &  various graph-based tasks    & 3D object recognition \& action recognition    \\
\cite{igor_reprogramming_2023}  &    Text        &  BERT       &  Protein sequence    & antibody sequence infilling with diversity   \\
\bottomrule
\end{tabular}
\end{adjustbox}
\caption{Summary of model reprogramming use cases. LSTM means long short-term memory, CNN/RNN means convolutional/recurrent neural network, API means application programming interface, and SOTA means state of the art. BERT stands for bidirectional encoder representations from transformers. GLUE stands for the general language understanding evaluation benchmark. GAN stands for generative adversarial network. CA stands for cellular automata. ViT stands for vision transformer. GNN stands for graph neural network.
The table is actively updated at \textcolor{blue}{\url{https://github.com/IBM/model-reprogramming}}.
}
\label{tab:reprog_comp}
%\vspace{-2mm}
\end{table*}

\paragraph{Output mapping layer.}
There are two major approaches to realizing the output mapping layer. The first approach is \textit{source-target label mapping}, by specifying a many-to-one surjective mapping $h: \cY_\cS \mapsto \cY_\cT$ from source class labels to target class labels. Take the example of reprogramming an ImageNet pre-trained source model for autism spectrum disorder (ASD) classification based on brain-region correlation graphs, the source label subset \{Tench,Goldfish,Hammerhead\} can be assigned to the target label \{ASD\}, and another non-overlapping source label subset can be assigned to the other target label \{non-ASD\}.
The prediction probability of a target class $t \in \cY_{\cT}$ on an input-transformed data sample $\xt_\cT$ in \eqref{eqn_input_transform} is the average prediction probability of the source labels specified by $h$ and $f_\cS$. Mathematically, let $\cB \subset \cY_{\cS}$ denote the subset of source labels mapping to the target label $t \in \cY_{\cT}$. Then, the class prediction of $t$ is the aggregated prediction based on $f_\cS$ over the assigned source labels, which is defined  as 
\begin{align}
  \label{eqn_prob}
  \textsf{Prob}(t|f_\cS(\xt_\cT)) = \frac{1}{|\cB|} \sum_{s \in \cB}  \textsf{Prob}(s|f_\cS(\xt_\cT))
\end{align}
where $|\cB|$ denotes the number of labels in $\cB$. In \cite{tsai2020transfer}, the authors show that frequency-based greedy label mapping based on original responses before training can improve reprogramming performance when compared with random label mapping. \cite{chen2023understanding} further shows that iterative greedy mapping gives better results.  Moreover, assigning more (but non-overlapping) source labels to a target label can also improve the final performance. Instead of using pre-specified label mapping, one can also learn a label mapping function $h$ by treating it as an optimal label assignment problem.

The second approach is \textit{adding a trainable dense layer} (linear head) with a set of trainable parameters $\omega$ between the source model's output of dimension $K_\cS$ (or the model's penultimate output) and the target model's output of dimension $K_\cT$, such as in \cite{kloberdanz2021improved,hambardzumyan2021warp,arifreprogrammable}.

\paragraph{Model training and evaluation.}
After introducing the two aforementioned modules, input transformation and output mapping layers, to a pre-trained source model $f_\cS$  (see Figure \ref{fig:reprog}),
the target-domain training set  $\{x_\cT^{(i)},y_\cT^{(i)}\}_{i=1}^n$ is used to evaluate the associated task loss $\textsf{Loss}(\yhat_\cT,y_\cT|\theta,\omega)$ (e.g., the cross entropy loss), where $\yhat_\cT=\textsf{Output-Mapping}(f_\cS(\textsf{Input-Transform}(x_\cT|\theta))|\omega)$ is the reprogrammed model prediction on the target-domain data sample $x_\cT$, and $y_\cT$ is the groundtruth target class label. The reprogramming parameters $\theta$ and $\omega$ ($\omega$ can be omitted if a source-target label mapping $h$ is used instead) are trained and updated based on $\textsf{Loss}(\yhat_\cT,y_\cT|\theta,\omega)$ and $\{x_\cT^{(i)},y_\cT^{(i)}\}_{i=1}^n$  in an end-to-end manner using an optimization algorithm such as a gradient-based method. In the restricted model access setting when the pre-trained source model $f_\cS$ is a black-box function that only provides model outputs at queried data inputs and back-propagation through $f_\cS$ is infeasible, such as a machine learning based application programming interface (API) or proprietary software, black-box model reprogramming can  be realized by using gradient-free methods (e.g., zeroth-order optimization \cite{liu2020primer}) for training  \cite{tsai2020transfer}. Finally, the optimized parameters $\theta^*$ and $\omega^*$ are used together with the pre-trained source model $f_\cS$ for evaluation.

\paragraph{Algorithmic procedure.}
Below we describe the generic algorithmic procedure for model reprogramming.
\begin{enumerate}[leftmargin=*]
    \item \underline{Initialization}: Load pre-trained source model $f_\cS(\cdot)$ and target domain training set $\{x_\cT^{(i)},y_\cT^{(i)}\}_{i=1}^n$; randomly initialize $\theta$ and $\omega$
    \item \underline{Input transformation}: Obtain transformed input data $\xt_{\cT}=\textsf{Input-Transform}(x_\cT|\theta)$, where $\theta$ is the set of trainable parameters for input transformation
    \item \underline{Output mapping}: Obtain the prediction on the target task via $\yhat_\cT=\textsf{Output-Mapping}(f_\cS(\xt_{\cT})|\omega)$, where $\omega$ is the set of trainable parameters for output mapping\footnote{In output mapping, trainable parameters may not be necessary if one uses a specified source-target label mapping function.}
    \item \underline{Model training}: Optimize $\theta$ and $\omega$ by evaluating a task-specific loss $\textsf{Loss}(\yhat_\cT,y_\cT|\theta,\omega)$ on 
    $\{x_\cT^{(i)},y_\cT^{(i)}\}_{i=1}^n$
    \item \underline{Outcome}: Reprogrammed model from $f_\cS(\cdot)$ with optimized trainable parameters $\theta^*$ and $\omega^*$ such that $\yhat_\cT=\textsf{Output-Mapping}(f_\cS(\textsf{Input-Transform}(x_\cT|\theta^*))|\omega^*)$
    
\end{enumerate}

\subsection{Model reprogramming use cases}
\label{sub_example}

Model reprogramming has shown success and improved performance for resource-efficient cross-domain machine learning on a wide range of data domains, pre-trained source models, and machine learning tasks. 
Table \ref{tab:reprog_comp} summarizes some studies on model reprogramming. Without loss of generality,  in what follows we highlight two representative use cases for each data format (continuous or discrete) featuring improved task performance and resource efficiency.

\paragraph{Continuous data domain}
\begin{itemize}[leftmargin=*]
    \item \textit{Black-box adversarial reprogramming (BAR)} \cite{tsai2020transfer}: BAR extends the original adversarial reprogramming framework \cite{elsayed2018adversarial} to enable reprogramming black-box models and demonstrates both data and cost efficiency in low-resource bio-medical applications. For example, the authors reprogram ImageNet pre-trained deep neural network classifiers to classify autism spectrum disorder (ASD) based on brain-region correlation graphs and report new state-of-the-art accuracy on this challenging data-limited task. Moreover, they also demonstrate that different commercial prediction APIs can be reprogrammed at an affordable cost to solve different tasks without knowing the details of the underlying machine learning model. With the cost of about 20 US dollars, two Clarifai.com APIs (Not Safe For Work and Moderation) and a traffic sign classification model trained by the Microsoft Custom Vision API were reprogrammed for several bio-medical classification tasks with good accuracy.
    \item \textit{Voice2Series (V2S)} \cite{yang2021voice2series}: V2S reprograms a speech model (acoustic signal as input and speech command prediction as output) for univariate time-series classification. Time series data include but are not limited to medical diagnosis (e.g., physiological signals such as electrocardiogram (ECG)), finance/weather forecasting, and industrial measurements (e.g., sensors and Internet of Things (IoT)). In general, time-series data are small-scale because they are not as abundant and easily accessible as speech data.
    Evaluated on UCR time series classification datasets \cite{dau2019ucr}, V2S outperforms or ties with the best baseline on 19 out of 30 datasets.
\end{itemize}

\paragraph{Discrete data domain}
\begin{itemize}[leftmargin=*]
\item \textit{Representation reprogramming via dictionary learning (R2DL)} 
\cite{vinod2023reprogramming}: Let $\cV_\cS \in \bbR^{N_\cS \times d}$ and $\cV_\cT \in \bbR^{N_\cT \times d}$ denote the vocabulary matrix of the source-domain and target-domain tokens, respectively, where the rows of $\cV_\cS$ and $\cV_\cT$ represent their token embedding vectors with the same dimension $d$ while their token numbers $N_\cS$ and $N_\cT$ can be different. R2DL transforms the embedding in $\cV_\cS$ obtained from a pre-trained language model to represent the embedding in $\cV_\cT$ by finding the parameters $\theta \in \bbR^{d\times d}$ such that $\cV_\cT \approx \cV_\cS \theta$, where a dictionary learning algorithm such as the K-SVD solver \cite{aharon2006k} is used
to obtain a column-wise sparse solution $\theta$. Then $\theta$ is further updated with the output mapping layer and a task-specific loss. In the reduced-data setting, the authors show improved performance over training from scratch when reprogramming pre-trained English language models (e.g., BERT and LSTM) for biochemical sequence classification including toxicity and antimicrobial peptide prediction. The same idea has been extended to protein sequence generation tasks \cite{igor_reprogramming_2023}. 

\item \textit{Word-level adversarial reprogramming (WARP)} \cite{hambardzumyan2021warp}: WARP inserts trainable prompt tokens to a pre-trained masked language model and uses a trainable dense layer for the output mapping. On the GLUE benchmark, WARP attains a comparable performance to modern language models while having a much smaller number of trainable parameters. The authors also demonstrate the data efficiency of WARP in the few-shot setting. 
%The results suggest that model reprogramming can be a resource-efficient approach
%for knowledge transfer from large pre-trained language models to downstream tasks.
\end{itemize}

\section{Theoretical Characterization of Model Reprogramming}
\label{sec_theory}

This section summarizes the theoretical characterization and interpretation of the working mechanism of model reprogramming based on current studies.

\subsection{Error analysis in representation alignment}

Using the notation in Table \ref{tab:notation}, let the source model $f_{\cS}$ be a pre-trained $K$-way neural network classifier $f_{\cS}(\cdot)=\eta(z_{\cS}(\cdot))$ with a softmax layer $\eta(\cdot)$ as the model output. Let $\ell(x,y) \triangleq \|f(x)-y\|_2$  denote the root mean squared error and let $\bbE_{\cD} [\ell (x,y)] \triangleq \bbE_{(x,y) \sim \cD} [\ell (x,y)]$ denote its population risk on $(x,y) \sim \cD$.
Under some mild assumptions such as one-to-one label mapping, \cite{yang2021voice2series} proves that the risk of the target task (target risk) via model reprogramming is upper bounded by the summation of two terms, the risk of the source model on the source task (source risk) and the representation alignment error measuring the distributional difference between the latent representations of the source and the reprogrammed target data based on the same source model $f_\cS$. The theorem is stated as follows.

\textbf{Theorem 1:} Let $\theta^*$ be the learned input transformation parameters from \eqref{eqn_input_transform} and assume one-to-one label mapping.
The population risk for the target task via reprogramming a $K$-way source neural network classifier $f_{\cS}(\cdot)=\eta(z_{\cS}(\cdot))$,  denoted by $\bbE_{\cD_{\cT}}[\ell_{\cT}(\xt_t(\theta^*),y_t)]$, is upper bounded by
\begin{align*}
&\underbrace{\bbE_{\cD_{\cT}}[\ell_{\cT}(\xt_t(\theta^*),y_t)]}_{\text{target~risk}} \leq \underbrace{\epsilon_{\cS}}_{\text{source~risk}}\\
&~+
     2\sqrt{K} \cdot \underbrace{\cW_1(\mu(z_{\cS}(\xt_t(\theta^*)),\mu(z_{\cS}(x_s)))_{x_t \sim \cD_{\cT},~x_s \sim \cD_{\cS}}}_{\text{representation~alignment~loss~via~reprogramming}},
\end{align*}
where $\xt_t(\theta^*)$ is an input-transformed target data sample defined in \eqref{eqn_input_transform} and $\cW_1(\mu_a,\mu_b)$ denotes  the Wasserstein-1 distance between two probability distributions $\mu_a$ and $\mu_b$.
%\\
%\textbf{Proof:} The proof follows \cite{yang2021voice2series}.

The theorem provides several insights into characterizing the success of model reprogramming, suggesting that the target risk via reprogramming can be improved (lower is better) when
(i) a better source model (in terms of lower source risk) is used, and (ii) the source and reprogrammed target data representations based on $f_\cS$ are better aligned (in terms of smaller Wasserstein distance). This analysis also explains that cross-domain model reprogramming is feasible as long as the target representations can be aligned with the source representations after reprogramming. In other words, model reprogramming can be interpreted as reusing a pre-trained source model as an efficient feature extractor to learn representation alignment.
Some numerical evidence is given in \cite{yang2021voice2series} to show that the Wasserstein distance indeed decreases during model reprogramming.

%\subsection{Representation alignment}
% tsne; SWD
% \begin{figure}[t]
% %\vspace{-2mm}
% \centering
% \includegraphics[width=0.48\textwidth]{ICMLWD_x.png}
% %\vspace{-6mm}
% \caption{\textcolor{red}{TBU} Training-time reprogramming analysis using V2S$_a$ and DistalPhalanxTW dataset~\cite{davis2013predictive}. All values are averaged over the test set. The rows are (a) validation (test) accuracy, (b) validation loss, and (c) sliced Wasserstein distance  (SWD)~\cite{kolouri2018sliced}.  }
% \label{fig:SWD}
% %\vspace{-2mm}
% \end{figure}

\subsection{Other interpretations}
In addition to representation alignment, we summarize different interpretations and analyses of model reprogramming based on existing works.
Using first-order approximation, \cite{zheng2021adversarial} shows that the optimal loss decrement in reprogramming is equivalent to the $\ell_1$-norm of the average input gradient. Consequently, model reprogramming is more successful when input gradients of target data are more aligned, and when inputs have higher dimensionality. \cite{hambardzumyan2021warp} motivates the data efficiency and fast adaptivity of model reprogramming in language models from the perspective of learning optimal trainable input prompts for efficient adaptation to downstream tasks. Notably, when the source and target domains are both in computer vision, model reprogramming essentially reduces to visual prompting \cite{bahng2022visual}.

\section{Open-ended Research Questions, Ongoing Efforts, and Opportunities}
\label{sec_discuss}

In this section, we discuss several open-ended research questions and opportunities for model reprogramming.

\paragraph{Model Reprogramming beyond Supervision.} Most of the existing works on model reprogramming focus on the supervised setting, where all data samples in target domains are associated with data labels for training. With the advance of semi-supervised (a mixture of labeled and unlabeled data samples), unsupervised (unlabeled data samples only), and self-supervised (self-generated pseudo labels) machine learning methods, it is essential to understand how unlabeled data and supervision-free training (possibly with data augmentation) improve model reprogramming.

\paragraph{Reprogramming Foundation Models.} The emergence of foundation models \cite{bommasani2021opportunities}, including large language models and generative AI applications, has led to a critical paradigm shift in the trend of machine learning research from designing task-specific and modality-dependent deep learning models to developing task-agnostic and modality-independent general-purpose models. Foundation models feature supervision-free pre-training and efficient finetuning to different downstream tasks. However, not every domain and every researcher has the luxury of accessing foundation models due to resource limitations. Demonstrating model reprogramming can serve as an affordable solution to repurposing and reusing a foundation model with resource efficiency can further accelerate AI-assisted scientific discovery and democratize machine learning research, as explored in \cite{xu2023towards}.

\paragraph{Reprogramming for Improved Model Properties.} Ensuring and instilling trustworthiness in machine learning based technology is the new norm for building a healthy and sustainable ecosystem between technology, end users, and our society. Studying how model reprogramming can be used as an efficient calibration tool to improve different trustworthiness properties while having minimal impact on the original utility is an important research direction. The model properties include fairness \cite{zhang2022fairness}, robustness \cite{chen2023visual}, explainability, privacy \cite{li2023exploring}, uncertainly quantification \cite{tang2022neural}, and energy efficiency \cite{sun2023neuralfuse}, to name few.

\paragraph{Joint Model Reprogramming and Fine-tuning.}
Theoretically, joint model reprogramming and fine-tuning on the right subset of the pre-trained model's parameters should perform no worse than standalone model reprogramming or transfer learning. However, in the resource-limited setting such joint training on a large-scale pre-trained model is challenging because identifying which model parameters to finetune is highly non-trivial. Studying how and when joint model reprogramming and fine-tuning can outperform standalone model reprogramming and transfer learning can lead to more resource-efficient machine learning. Notably, when applied to the task of low-resource speech commands recognition \cite{yen2021study} and music genre classification \cite{hung2023low}, such joint training with careful fine-tuning is shown to deliver improved performance. Moreover, although originally not cast as a model reprogramming method, the proposal of universal compute engine \cite{lu2021pretrained} that reuses a pre-trained transformer for attaining non-trivial performance on a diverse set of tasks, by joint fine-tuning the input embedding, the layer-norm parameters, and training the output mapping layer, suggests great potential for resource-efficient machine learning.

%\paragraph{Input-aware Model Reprogramming.} Many model reprogramming methods are based on learning a universal input transformation function. To fully understand the capability of model reprogramming, further investigation on the design of input-aware model reprogramming is necessary.

\paragraph{Implications on Machine Learning Systems with Heterogeneous Computing.} The advantage of ``no model fine-tuning'' and resource efficiency in model reprogramming can be favorable to machine learning systems featuring heterogeneous computation and memory constraints, such as edge computing, cloud computing, and federated learning. For energy and memory efficiency, the source model (which is usually large) can be intact and stored on the server side, while the input transformation and output mapping layers can be implemented at the edge device for reprogramming. During training, the communication cost between the server and an edge device is expected to be greatly reduced because such a paradigm does not require fine-tuning the model parameters at the server. It also spares the need for the edge device to download and store a local copy of the model from the server. \cite{arifreprogrammable} improves the privacy-utility tradeoff in differentially private training with model reprogramming over standard transfer learning methods in both centralized and federated learning scenarios.

\bibliography{adversarial_learning}

\end{document}

%% file: CPY_notation_V1.tex
\newcommand{\yhat}{\widehat{y}}

\newcommand{\xt}{\widetilde{x}}

\newcommand{\cB}{\mathcal{B}}
\newcommand{\cS}{\mathcal{S}}
\newcommand{\cD}{\mathcal{D}}

\newcommand{\cW}{\mathcal{W}}

\newcommand{\cX}{\mathcal{X}}
\newcommand{\cY}{\mathcal{Y}}

\newcommand{\cV}{\mathcal{V}}

\newcommand{\cT}{\mathcal{T}}

\newcommand{\bbR}{\mathbb{R}}
\newcommand{\bbE}{\mathbb{E}}